\pdfoutput=1
\documentclass[11pt]{article}

\usepackage{ACL2023}
\usepackage{tablefootnote}

\usepackage[defaultlines=2,all]{nowidow}

\usepackage{times}
\usepackage{latexsym}
\usepackage{graphicx}
\usepackage{algorithm}
\usepackage{algpseudocode}
\algrenewcommand\algorithmicindent{0.85em}%
\usepackage[most]{tcolorbox}
\usepackage{float}
\usepackage{placeins}
\usepackage{makecell}

\usepackage[T1]{fontenc}
\usepackage[utf8]{inputenc}

\usepackage{microtype}

\usepackage{inconsolata}

\usepackage{tabularx}
\usepackage{xspace}

\usepackage{booktabs}
\usepackage{multirow}

\usepackage[table]{xcolor}

\usepackage{amsmath,amsfonts,amssymb}

\usepackage{enumitem}

\usepackage{amssymb}% http://ctan.org/pkg/amssymb
\usepackage{pifont}% http://ctan.org/pkg/pifont
\definecolor{darkerred}{RGB}{190, 0, 0} % RGB model

\newcommand{\cmark}{\textcolor{teal}{\ding{51}}}%
\newcommand{\xmark}{\textcolor{darkerred}{\ding{55}}}%

\newcommand{\OURS}{\textsc{LLM2LLM}\xspace}
\newcommand{\TEACHER}{{\mathcal{M}_\text{teacher}}}

\newcommand{\graycell}[1]{\cellcolor{teal!10}#1}

\newcommand{\tref}[1]{Table~\ref{#1}}
\newcommand{\sref}[1]{Section~\ref{#1}}
\newcommand{\fref}[1]{Figure~\ref{#1}}

\newcommand\blfootnote[1]{%
  \begingroup
  \renewcommand\thefootnote{}\footnote{#1}%
  \addtocounter{footnote}{-1}%
  \endgroup
}

\title{\OURS: Boosting LLMs with Novel Iterative Data Enhancement}

\author{Nicholas Lee$^{*1}$ \enskip\enskip Thanakul Wattanawong$^{*1}$ \enskip\enskip Sehoon Kim$^{1}$\\ \bf Karttikeya Mangalam$^{1}$ \enskip\enskip Sheng Shen$^{1}$ \enskip\enskip Gopala Anumanchipalli$^{1}$\\ \bf Michael W. Mahoney$^{1,2,3}$ \enskip\enskip Kurt Keutzer$^{1}$ \enskip\enskip Amir Gholami$^{1,2}$\\
$^{1}$UC Berkeley\qquad$^{2}$ICSI\qquad$^{3}$LBNL \\
{{\small{\{nicholas.lee, j.wat, sehoonkim, mangalam, s.sheng,  gopala, mahoneymw, keutzer, amirgh\}@berkeley.edu}}}
}

\begin{document}
\maketitle
\begin{abstract}
Pretrained large language models (LLMs) are currently state-of-the-art for solving the vast majority of natural language processing tasks.
While many real-world applications still require fine-tuning to reach satisfactory levels of performance, many of them are in the low-data regime, making fine-tuning challenging.
To address this, we propose \OURS, a targeted and iterative data augmentation strategy that uses a teacher LLM to enhance a small seed dataset by augmenting additional data that can be used for fine-tuning on a specific task.
\OURS (1) fine-tunes a baseline student LLM on the initial seed data,
(2) evaluates and extracts data points that the model gets wrong, 
and (3) uses a teacher LLM to generate synthetic data based on these incorrect data points, which are then added back into the training data.
This approach amplifies the signal from incorrectly predicted data points by the LLM during training and reintegrates them into the dataset to focus on more challenging examples for the LLM.
Our results show that \OURS significantly enhances the performance of LLMs in the low-data regime, outperforming both traditional fine-tuning and other data augmentation baselines.
\OURS reduces the dependence on labor-intensive data curation and paves the way for more scalable and performant LLM solutions, allowing us to tackle data-constrained domains and tasks.
We achieve improvements up to 24.2\% on the GSM8K dataset, 32.6\% on CaseHOLD, 32.0\% on SNIPS, 52.6\% on TREC and 39.8\% on SST-2 over regular fine-tuning in the low-data regime using a Llama-2-7B student model.
Our code is available at~\url{https://github.com/SqueezeAILab/LLM2LLM}. 
\end{abstract}
\section{Introduction}
\blfootnote{*Equal contribution}
Pretrained large language models (LLMs) have achieved impressive performance on various benchmarks and datasets that have previously required specialized neural network architectures.
For many of these general benchmarks \cite{hendrycks2020measuring, zhong2023agieval}, LLMs are prompted with custom instructions or in-context examples.

However, in various real-world applications, these prompting strategies are not a one-size-fits-all solution.  
For instance, LLMs have a limit on the amount of input context they can process, thus limiting the number of in-context examples or instructions we can input to make the LLM follow a certain behavior. 
For simple tasks that are closely aligned with the data that the LLM was pretrained on, extensive prompting may not be necessary.
However, applying LLMs to specialized domains (e.g., a specific medical field \cite{nori2023can} or private data with niche protocols) can be more challenging, often requiring prohibitively long prompts to achieve adequate performance. 
Even if the prompt length does not exceed the limit, processing long prompts increases the latency and cost of each inference. 
Additionally, LLMs also tend to forget or ignore information in long contexts~\cite{liu2023lost}, leading to potential accuracy drops even when the model can handle long input prompts.
While Retrieval-Augmented Generation (RAG) \cite{lewis2020retrieval} has been developed
to address some of these challenges, it may sometimes retrieve irrelevant passages or documents, which can potentially degrade the generation performance. 
Furthermore, RAG does not necessarily solve the latency and cost issue as 
processing a long input prompt may still be required.

A promising method for addressing this is fine-tuning.
With the emergence of Parameter Efficient Fine-tuning (PEFT)~\cite{hu2021lora, peft}, the computational resources required to fine-tune a task-specific LLM have decreased significantly.
However, herein lies a new problem: successful fine-tuning requires enough training data.
This can be challenging for some applications, where we only have access to a small amount of task-specific data.
Often, collecting, cleaning, and labeling additional data can be costly and time-consuming. 
So the key question is:
\textit{how should we increase the user's training data to be enough for fine-tuning?}

Data augmentation is a known method that could help effectively expand the training dataset. 
For natural language processing (NLP) tasks, one can use approaches such as synonym replacement,
character replacement (e.g., by intentionally introducing spelling errors), random swapping, and
back translation, just to name a few~\cite{wei2019eda, belinkov2017synthetic, coulombe2018text, zhang2018mixup}. 
However, these approaches fail to effectively expand the training data for fine-tuning LLMs in the case of new and specialized tasks, as we will show later in \sref{sec:baseline_results}.

To address this, several recent papers have explored using an LLM to expand
the fine-tuning dataset~\cite{dai2023auggpt, kumar2020data, zhou2023lima,chen2023alpagasus, cao2023instruction, wei2023instructiongpt4, zhu2023minigpt4}. This approach has
proven to be more effective than traditional data augmentation methods.
However, these approaches often apply LLM-based data augmentation on all of the
available training dataset, without considering the LLM's prediction accuracy on individual
training data points. We have observed that for various reasoning tasks such as arithmetic and reading comprehension, the LLM correctly solves simpler examples in the fine-tuning dataset, but may struggle with harder examples. 
It will be sub-optimal to keep augmenting data points for which the LLM is already achieving high accuracy on.

To address these challenges, we introduce \OURS, a new targeted and iterative data augmentation framework that uses a teacher LLM to expand the training dataset, with a targeted and iterative approach. 
In more detail, we make the following contributions:

\begin{itemize}[leftmargin=3mm]
    \vspace{-1mm}
    \item 
    We propose \OURS, a targeted and iterative LLM-based data augmentation technique that efficiently and effectively augments small task-specific datasets. 
    \OURS achieves this by (1) fine-tuning a student LLM on the initial dataset, (2) evaluating on the training data and extracting data points which the model got incorrect after training, and (3) using a Self-Instruct \cite{wang-etal-2023-self-instruct} style data augmentation to augment these data points, which are then added back into the training data (\sref{sec:llm2llm}).
    \vspace{-1mm}
    \item 
    We benchmark \OURS on randomly sampled subsets of GSM8K \cite{cobbe2021gsm8k}, CaseHOLD \cite{zhengguha2021}, SNIPS \cite{coucke2018snips}, TREC \cite{li-roth-2002-learning} and SST-2 \cite{socher2013recursive} in order to evaluate the effectiveness of our approach in the low-data regime (\sref{sec:main_results}). 
    Here, we get up to a 24.2\% improvement on GSM8K, 32.6\% on CaseHOLD, 32.0\% on SNIPS, 52.6\% on TREC, and 39.8\% on SST-2 (\tref{table:results}).
    \vspace{-1mm}
    \item 
    We conduct a series of ablations studies comparing \OURS to several existing baselines as well as to variants of \OURS to evaluate the effectiveness of our design decisions (\sref{sec:ablations}).
    We observe that both the iterative and targeted nature of \OURS are critical to improving model performance.
\end{itemize}
\section{Background and Related Work}
\label{sec:related_work}

\subsection{Instruction Following LLMs}

The earliest works \cite{wei2021finetuned, longpre2023flan, chung2022scaling, aribandi2021ext5, sanh2021multitask, muennighoff2022crosslingual, wang2022knowda, mishra2022cross,wang2022super,xu2022zeroprompt} in instruction fine-tuning involved gathering and processing different existing NLP datasets in order to improve the performance of LLMs on a wide range of tasks.
Self-Instruct~\cite{wang-etal-2023-self-instruct} removed the reliance on existing datasets by introducing a framework for bootstrapping instruction datasets with the outputs of the model itself. 
Follow-up work \cite{ouyang2022training, taori2023alpaca, koala_blogpost_2023, vicuna2023, xu2023wizardlm, mukherjee2023orca, mitra2023orca, kang2023knowledgeaugmented, nori2023can} took advantage of stronger models \cite{achiam2023gpt, touvron2023llama, touvron2023llama2} in order to fine-tune stronger general-purpose instruction-following models.

%%%%%%%%%%%%%%%%%%%%%%%%%%%%%%%%%%%%%%%%%%%%%%%%%%%%%%%%%%%%%%%%%%%%%%%%%
\begin{figure*}[!t]
  \centering
  \vspace{-2mm}
  \includegraphics[width=\textwidth]{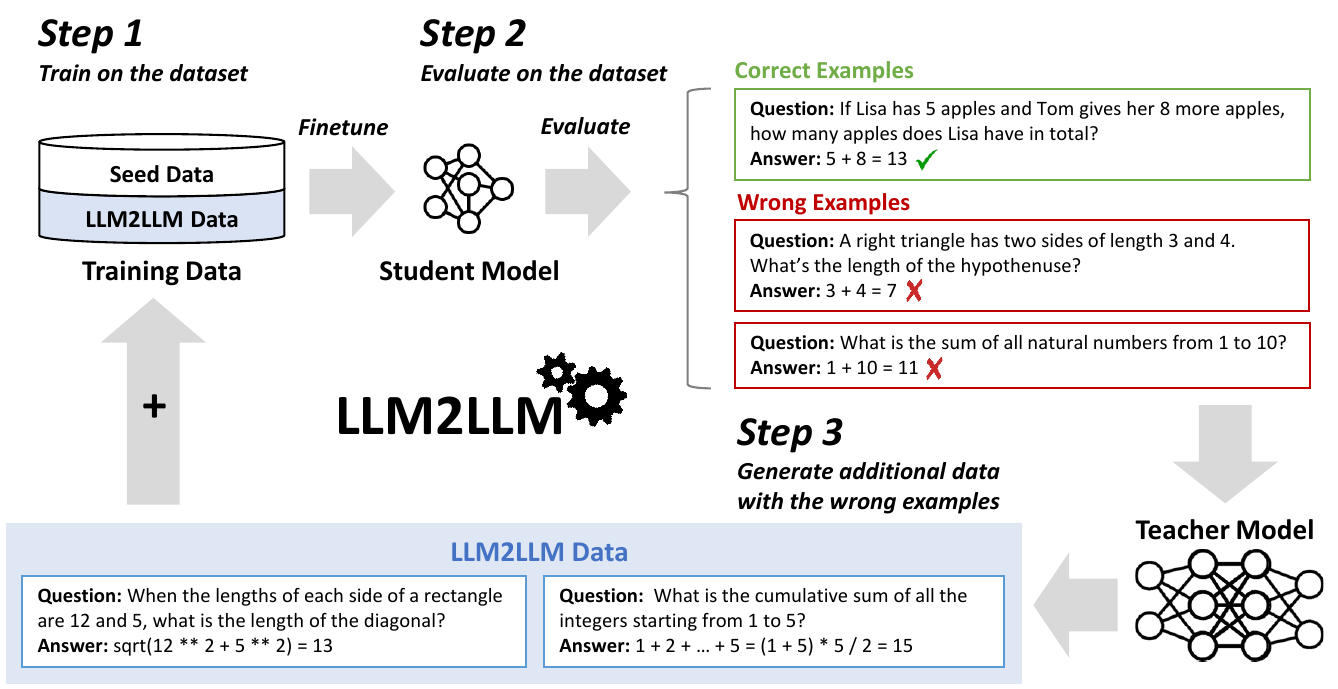} 
  \vspace*{-7mm}
  \caption{\OURS: Boosting LLMs with Novel Iterative Data Enhancement. One iteration of \OURS begins with training and evaluating the model on the training data. Incorrect answers from the training data are used as inputs to generate extra samples with similar styles to the teacher model. Then, a new student model is trained using a combination of the old training data and newly generated samples.
  After the model is fine-tuned, we evaluate and find questions that the model got incorrect.
  The teacher model is used to generate additional data points based on the wrong examples, which test for similar concepts and ideas.
  These synthetic data points are folded back into training dataset.
  This process then repeats, training the student model on increasingly targeted data points.}
  \label{fig:autollm}
  \vspace{-4mm}
\end{figure*}
%%%%%%%%%%%%%%%%%%%%%%%%%%%%%%%%%%%%%%%%%%%%%%%%%%%%%%%%%%%%%%%%%%%%%%%%%

\subsection{Self-Improving LLMs}

Various early works \cite{zelikman2023selftaught, haluptzok2023language, zelikman2022star, madaan2023self, gulcehre2023reinforced, singh2023beyond} explore using self-improvement for fine-tuning LLMs.
These works generally filtered the outputs of the model before fine-tuning it on its own outputs.
\OURS differs from these methods, as we do not directly fine-tune on the outputs of our own model, and we employ a teacher model to provide feedback in the form of synthetic data.

Concurrent with our work, several papers have been recently published that use an iterative approach to improving LLMs \cite{chen2024self, anil2023palm, burns2023weaktostrong, li2023self, yuan2024self}.
These works combine ideas from Reinforcement Learning (RL) and Self-Play \cite{samuel2000some, tesauro1995temporal} in order to iteratively build stronger LLMs by fine-tuning on the outputs of the model itself.
\OURS is distinguished by how it focuses on the low-data regime for task-specific fine-tuning of LLMs whereas others are attempting to create stronger general-purpose LLMs.  
In addition, our technique exclusively uses the data points that the model got incorrect during training, and it uses the teacher model to augment these data points. 
Instead of providing feedback in the form of a critique or rationale, the teacher model's feedback is only in the form of synthetic data points, which simplifies the training pipeline.

\subsection{Data Augmentation}
Data augmentation has been long studied in NLP.
Early work augmented at the character \cite{belinkov2017synthetic, coulombe2018text} and word \cite{wei2019eda} level.
Notably, Easy Data Augmentation (EDA) \cite{wei2019eda} was a popular early method that used word level augmentations: synonym replacement, random insertion, swap, and deletion to augment data for text classification. 
We refer the reader to~\cite{feng-etal-2021-survey} for a more complete summary of data augmentation in NLP.

A popular new approach is to use LLMs themselves to synthesize new training data \cite{deng2023rephrase, prasad2023rephrase, fu2023specializing, dai2023auggpt, ubani2023zeroshotdataaug, 
fang2023using, liu2023tinygsm, yu2023metamath, kumar2020data, yoo2021gpt3mix, wang2021want, ding2023gpt, li2023coannotating, liang2023let}.
A noteworthy example is AugGPT \cite{dai2023auggpt}, which used ChatGPT to rephrase text to augment text classification tasks.

Many of these techniques generate very large amounts of synthetic data.
Recent work \cite{chen2023alpagasus, cao2023instruction, wei2023instructiongpt4, zhou2023lima} found that one could replicate the results of fine-tuning on these large datasets with significantly smaller subsets.

\section{Methodology}

We assume that we are given an LLM model $\mathcal{M}$ (e.g., GPT-3.5 or Llama-2-7B) that is pre-trained on some source dataset (e.g., Common Crawl).
The goal is to adapt $\mathcal{M}$ (hereon called the student model) to a new target domain by using a small seed dataset $D$, where $D$ potentially has unseen characteristics, compared to the pre-trained dataset (e.g., a medical dataset with specific terminology, or a private database with specific characteristics).
In this case, the model's zero-shot or fine-tuned performance is likely to be unsatisfactory.
While strategies to address this challenge have been explored, e.g., through enhanced few-shot learning methods as discussed in Section \ref{sec:related_work}, here we strictly focus on enriching the provided target
dataset $D$ with an LLM.
This method is orthogonal to the aforementioned techniques, offering a complementary solution that can be applied alongside them.

To enrich $D$, AugGPT~\cite{dai2023auggpt} has introduced a promising approach that generates additional augmented data by applying a prompted LLM to all available data points in the target training dataset. 
However, this method falls short by indiscriminately augmenting data without considering the student model's varying performance across different data points.
For instance, the model may easily solve the majority of the dataset, but it may struggle with a small subset of more challenging examples. 
In this case, rather than indiscriminately expanding the dataset by replicating simpler cases, a better augmentation strategy would be to generate more data points that align conceptually with these challenging examples.
This is because the former approach could lead to longer training time without noticeable performance improvement.

%%%%%%%%%%%%%%%%%%%%%%%%%%%%%%%%%%%%%%%%%%%%%%%%%%%%%%%%%%%%%%%%
\begin{algorithm}[t!]
\footnotesize
\caption{\OURS: Boosting LLMs with Novel Iterative Data Enhancement. Given a seed dataset $D^0$, we finetune the model $\mathcal{M}_\text{student}^i$, evaluate, and extract training set data points that the model gets wrong. These are used to generate new training data points using the teacher model $\TEACHER$ for the next step.}
\begin{algorithmic}[1]
\Procedure{\OURS}{${\mathcal{M}_\text{student}^0}$, $\TEACHER$, $D^0$}
\State $i \gets 0$
\While{$i < n$}
    \State $\mathcal{M}_\text{student}^i \gets$ Finetune($\mathcal{M}_\text{student}^0$, $D^{i}$)
    \State $E^i \gets$ Evaluate($\mathcal{M}_\text{student}^i$, $D^0$) {\scriptsize\textcolor{gray}{\algorithmiccomment{Evaluate on seed data}}}
    \State $W^i \gets$ Filter($E^i$,$D^0$) {\scriptsize\textcolor{gray}{\algorithmiccomment{Keep wrong answers}}}
    \State $A^{i} \gets$ Generate($\TEACHER$, $W^i$)  {\scriptsize\textcolor{gray}{\algorithmiccomment{Augment using teacher}}}
    \State $D^{i+1} \gets$ $D^i + A^i$ {\scriptsize\textcolor{gray}{\algorithmiccomment{Append to data}}}
    \State $i \gets i+1$
\EndWhile
\State Evaluate $M_\text{student}^*$
\EndProcedure
\end{algorithmic}
\label{alg:autollm}
\end{algorithm}
%%%%%%%%%%%%%%%%%%%%%%%%%%%%%%%%%%%%%%%%%%%%%%%%%%%%%%%%%%%%%%%%

Here, we propose a more general formulation of an LLM-based data augmentation pipeline that addresses the aforementioned limitation.
To do so, we consider the following iterative process:
\begin{equation}
\label{eqn:general_formula}
\small
    D^{n+1} = f(\TEACHER, \mathcal{M}_\text{student}, D^n, \cdots, D^0).
\end{equation}
In Equation~\eqref{eqn:general_formula}, $\TEACHER$ is the teacher model, $\mathcal{M}_\text{student}$ is the student model (potentially being fine-tuned in many iterations), $n$ refers to the $n^{th}$ step of data augmentation, $D^{n+1}$ is 
the new training dataset at the next iteration, and $f$ is the data-generation algorithm.
At each step, the teacher model has access to how the student model performs at the $n^{th}$ step (e.g., correct/incorrect labels, or possibly prediction distributions for white-box models), and based
on that it can edit training data points for the next iteration.

Note that \OURS is different from knowledge distillation~\cite{hinton2015distilling}.
Knowledge distillation is generally applicable to cases where the teacher model has high accuracy on the target
data. 
In contrast, in this case, it is possible that the teacher model also performs sub-optimally on the
target data (e.g., in the private database case, where the teacher lacks domain-specific knowledge). However, if the teacher model has enough reasoning capability
to produce conceptually similar but semantically different examples when it is given both the prompt and answer, then our framework can improve performance.

In \OURS, we consider a specific instantiation of Equation~\eqref{eqn:general_formula}, as discussed next.

% ------------------------------------------------
\subsection{\OURS}
\label{sec:llm2llm}

%%%%%%%%%%%%%%%%%%%%%%%%%%%%%%%%%%%%%%%%%%%%%%%%%%%%%%%%%%%%%%%%%%%%%%%%%
\begin{table*}[h!]
\centering
\footnotesize{
\setlength{\tabcolsep}{6pt}{
\begin{tabular}{c|c|c|c|c>{\bfseries}c}
\toprule
\multirow{2}{*}{\raisebox{-0.5ex}{\textbf{Dataset}}} & \multirow{2}{*}{\raisebox{-0.5ex}{\textbf{\% Data}}} & \multirow{2}{*}{\raisebox{-0.5ex}{\textbf{\# Seed Examples}}} & \multirow{2}{*}{\raisebox{-0.5ex}{\textbf{\# Augmented}}}  & \multicolumn{2}{c}{\textbf{Test Accuracy (\%)}}  \\
\cmidrule{5-6}
& & & & \textbf{Baseline} & \textbf{\OURS} \\
\midrule
 \multirow{7}{*}{GSM8K} & 0 & 0 & 0 & 0.00\footnotemark & \graycell{N/A} \\
 & 1 & 74 & 391 & 0.99 & \graycell{19.56}  \\
& 2 & 149 & 802 & 1.52 & \graycell{25.70}   \\
&  5 & 373 & 1641 & 9.63 & \graycell{27.07}  \\
& 10 & 747 & 2573 & 21.27& \graycell{30.93}  \\
&  20 & 1494 & 4028 & 25.70 & \graycell{35.03}  \\
& 50 & 3737 & 8252 & 33.89 & \graycell{38.67} \\
&  100 & 7473 & 14925 & 36.01 & \graycell{41.24} \\

\midrule
 \multirow{9}{*}{CaseHOLD}
 & 0 & 0 & 0 & 12.28 & \graycell{N/A} \\
 & 0.5 & 225 & 490 & 33.94 & \graycell{66.50} \\
& 1 & 450 & 751 & 46.25 & \graycell{70.97} \\
& 2 & 900 & 580 & 69.44 & \graycell{74.97}\\
& 5 & 2250 & 423 & 74.14 & \graycell{76.83} \\
& 10 & 4500 & 505 & 77.03 & \graycell{78.21} \\
& 20 & 9000 & 1100 & 78.00 & \graycell{78.97} \\
&  50 & 22500 & 2709 & 80.39 & \graycell{82.92} \\
& 100 & 45000 & 5805 & 87.94 & \graycell{88.14} \\
\midrule
\multirow{4}{*}{SNIPS} & 0 & 0 & 0 & 11.86 & \graycell{N/A} \\
& 0.5 & 70 & 38 & 60.14 & \graycell{92.14} \\
& 0.8 & 105 & 109 & 69.71 & \graycell{93.71} \\
& 1.0 & 140 & 91 & 85.43 & \graycell{93.86} \\

\midrule
\multirow{4}{*}{TREC} & 0& 0 & 0 & 11.20 & \graycell{N/A} \\
& 1.1 & 60 & 105 & 26.20 & \graycell{78.80} \\
& 1.6 & 90 & 22 & 80.80 & \graycell{90.20} \\
& 2.2 & 120 & 44 & 81.20 & \graycell{91.20} \\

\midrule
\multirow{4}{*}{SST-2\footnotemark} & 0 & 0 & 0 & 27.06 & \graycell{N/A} \\
& 0.02 & 20 & 44  & 52.87 & \graycell{92.66} \\
& 0.04 & 30 & 46 & 62.04 & \graycell{93.00} \\
& 0.06 & 40 & 14 & 82.80 & \graycell{94.04} \\

\bottomrule
\end{tabular}
}
}
\vspace{-2mm}
\caption{\OURS on datasets under evaluation. The \textit{\% Data} and \textit{\# Seed Examples} columns indicate the percentage and number of data points respectively that were sampled from the original training data as seed data.
The \textit{\# Augmented} column shows the number of data points created by \OURS. The last column (\textit{Test Accuracy \%}) shows the baseline accuracy from fine-tuning with the original seed examples (Baseline), as well as when training with augmented data added to the dataset (\OURS). Overall, test accuracy improves significantly with \OURS, especially in low data regimes.}
\label{table:results}
\vspace{-5mm}
\end{table*}
%%%%%%%%%%%%%%%%%%%%%%%%%%%%%%%%%%%%%%%%%%%%%%%%%%%%%%%%%%%%%%%%%%%%%%%%%

The end-to-end algorithm of \OURS is presented in Algorithm~\ref{alg:autollm}.
Inspired by Self-Instruct \cite{wang-etal-2023-self-instruct}, we use the teacher model $\TEACHER$ to generate synthetic data from the data points that the model got incorrect during training in order to target these deficiencies in the student model. In more detail, we first train the baseline student model $\mathcal{M}_\text{student}$ on the provided target data $D^0$, and we evaluate its performance (lines 4-5 of Algorithm~\ref{alg:autollm}). 
We then filter the results and keep the incorrect training examples
that the student model struggled to answer correctly ($E^i$ in line 6). Then the teacher model is prompted
to create additional training data points that are conceptually aligned but semantically different (line 7, see \sref{sec:teacher_prompting} for specifics on the prompt).
The teacher model does not necessarily need to be bigger, although that could potentially improve performance.
The primary requirement for the teacher model is to have reasoning capability to be able to follow the
data augmentation instruction, and the ability to create data points similar to the incorrect examples.
This process is schematically illustrated in~\fref{fig:autollm}.

%%%%%%%%%%%%%%%%%%%%%%%%%%%%%%%%%%%%%%%%%%%%%%%%%%%%%%%%%%%%%%%%%%%%%%%%%
\begin{figure*}[htbp]
    \centering
    \vspace*{-7mm}
    \includegraphics[width=\textwidth]{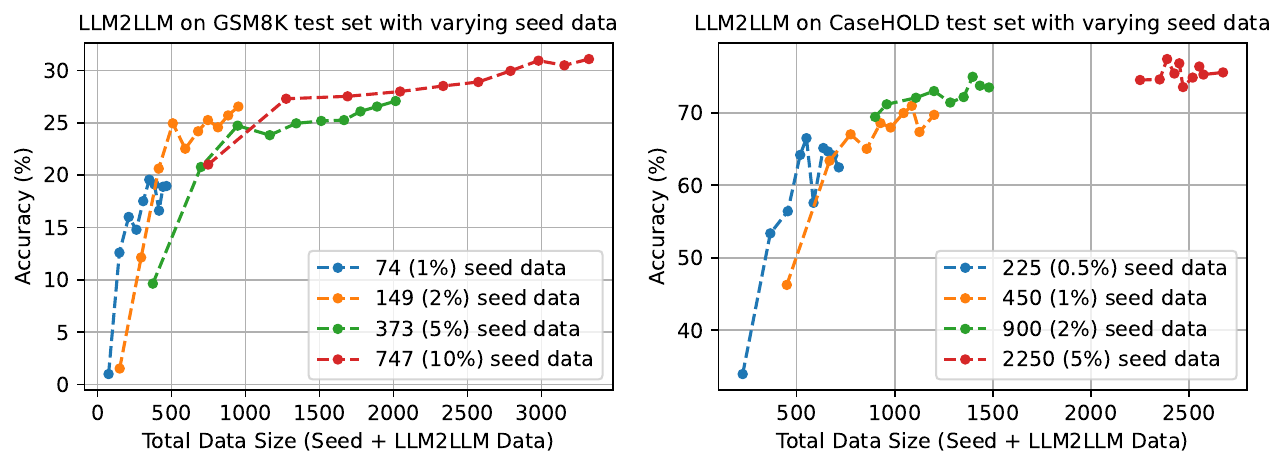}
    \caption{
    \OURS on GSM8K (left) and CaseHOLD (right) with various seed data sizes.
    Each line shows the test accuracy of the finetuned Llama-2-7B model with each step of \OURS with varying seed dataset size. The first (left-most) data point on each line represents finetuning only on the seed data. 
    Each point afterward corresponds to the performance after one more iteration of \OURS.
    The total data size (x-axis) represents the total amount of seed plus \OURS data that was used to train the model at that step. 
    By applying \OURS with low amounts of seed data and iteratively improving the training dataset, we can attain significant performance improvements.
    In particular, we can see that running \OURS can match or even exceed the performance of simply annotating more real data in some cases (detailed breakdown provided in \tref{table:results}). 
    }
    \label{fig:all_datasets_results_grid}
    \vspace{-3mm}
\end{figure*}
%%%%%%%%%%%%%%%%%%%%%%%%%%%%%%%%%%%%%%%%%%%%%%%%%%%%%%%%%%%%%%%%%%%%%%%%%

\footnotetext[1]{The 0\% here is because our prompting does not include the formatting that we used to extract the answer with. Adding some instructions to make sure the output format is correct results in a 0.30\% test accuracy, which is in line with the rest of the results in \tref{table:results}.}
\footnotetext[2]{SST-2 has no test set, therefore we evaluate on the dev set instead, see \sref{sec:baselines}.}

A subtle but important design decision in \OURS is that we only use examples from the seed data when prompting the teacher model to generate additional data points.
This is similar to Alpaca~\cite{taori2023alpaca}, but unlike Evol-Instruct~\cite{xu2023wizardlm}.
There are two main reasons for this.
First, our approach prevents data degradation from multiple augmentation iterations.
Early experiments revealed that while the teacher model could generate high-quality augmentations, some examples contained logical errors.
Therefore, further augmentation applied to these examples could potentially propagate the error, degrading the quality of the dataset over time.
This is highlighted in our ablation studies in \tref{table:aug_seed_data}, where using both seed and synthetic data for data augmentation leads to an accuracy drop.

Second, this approach limits the amount of new data being generated overall. 
Suppose that the original seed dataset is of size $n$, and at each iteration, the student model gets $p_i$ proportion of the training dataset $D^i$ wrong, where $0 < p_i < 1$.
If we include the augmented data into the seed data for data generation, then the size of the dataset $D^j$ at step $j$ will be 
\begin{equation*}
    |D^j| = n \prod_{i=0}^j (1 + p_i) \geq n (1 + p_{\text{min}})^j .
\end{equation*}
This has a lower bound that grows exponentially with each step. 
Limiting the input wrong answers $W^i$ during dataset generation to only data from the original seed data allows us to bound the total number of training data points to 
\begin{equation*}
    |D^j| = n + \sum_{i=0}^j np_i \leq n(1 +j p_{\text{max}}) ,
\end{equation*}
which has an upper-bound that grows linearly with the number of steps.
The empirical evaluations shown in~\sref{sec:adaptation_ablation} (\tref{table:aug_seed_data}) corroborates this.

\section{Results}

\subsection{Experimental Setup}
\label{sec:experimental_setup}
To evaluate the performance of \OURS, we applied our framework to fine-tune Llama-2-7B on various target datasets, including GSM8K \cite{cobbe2021gsm8k}, CaseHOLD \cite{nori2023can}, SNIPS \cite{coucke2018snips}, TREC \cite{li-roth-2002-learning} and SST-2 \cite{socher2013recursive}.
We subsampled these datasets with different sampling rates from 0.02\% to 50\% to evaluate performance across different low-data regimes. Our teacher model for these results is GPT-3.5 (1106 release) unless otherwise specified. We considered several other teacher models, including GPT-3.5, GPT-4-Turbo, Llama-2-70B~\cite{touvron2023llama2}, and Airoboros-L2-70B~\cite{airoboros} in~\sref{sec:teacher_ablation}.
We include a more detailed experimental setup in~\sref{sec:experimental_setup_appendix}.
Additionally, we conducted additional experiments (\sref{sec:word_overlap}) to ensure that the augmented data from the teacher models differs from the test dataset used to evaluate final model accuracy.
This addresses the issue of potential test data leakage that could have happened if the teacher model had been trained on similar data.

% -------------------------------------------------------------------
\subsection{Main Results}
\label{sec:main_results}
Here, we discuss \OURS's performance with varying amount of training data by presenting results for fine-tuning Llama-2-7B on GSM8K using GPT-3.5 as the teacher model.
We then discuss how these trends extend to different datasets (\tref{table:results}).

The final model accuracy after applying 10 iterations of \OURS is given in~\tref{table:results}.
For a low-data regime with 74 available examples (i.e., 1\% of the GSM8K training dataset), 
vanilla fine-tuning achieves only 0.99\% test accuracy.
However, \OURS  boosts the accuracy to 19.56\% by iteratively generating 391 additional examples based on data points where the model makes mistakes.
With slightly more available data of 149 seed examples (i.e., 2\% of the training dataset) we can
achieve 25.70\% accuracy. 
As shown in the baseline accuracy with 20\% data in \tref{table:baseline}, we would need over 10$\times$ more training data points to match this accuracy if we only rely on vanilla fine-tuning.
We also highlight that \OURS can lead to noticeable gains with data-sufficient regimes (e.g., 100\% data), albeit at a smaller improvement over the baseline compared to lower-data regimes.

We observe a similar trend for CaseHOLD, SNIPS, TREC, and SST-2, where \OURS helps improve performance in the low-data regime.
Interestingly, \OURS generally generates proportionally more augmented data for GSM8K than other datasets. This is because the baseline accuracy is lower for GSM8K overall, suggesting that it is a more difficult dataset compared to the others. However, in all cases, we find that \OURS helps recover a high-performing model.

In~\fref{fig:all_datasets_results_grid}, we also illustrate how the baseline accuracy improves on GSM8K and CaseHOLD with each iteration of applying \OURS. 
We can observe a rapid increase in test accuracy in the first few iterations of \OURS, especially in lower-data regimes. 

%%%%%%%%%%%%%%%%%%%%%%%%%%%%%%%%%%%%%%%%%%%%%%%%%%%%%%%%%%%%%%%%%%%%%%
\begin{table}[t!]
\centering
\footnotesize{
\setlength{\tabcolsep}{3.5pt}{
\begin{tabular}{c|c|c|c|c}
\toprule
\textbf{Dataset} & \textbf{Technique} & \textbf{\# Seed}  & \textbf{Total Aug.} & \textbf{Acc. (\%)} \\
\midrule
\multirow{5}{*}{GSM8K} &   Fine-tuning & \multirow{5}{*}{100} & 0  & 1.59 \\
                             & EDA & & 500 & 15.16 \\
                             & AugGPT && 500  & 18.12 \\
                             &  More Data && 471  & 19.86 \\
                              & \graycell{\OURS} & \graycell& \graycell{471} & \graycell{\textbf{23.73}} \\
\midrule
\multirow{5}{*}{CaseHOLD}  &  Fine-tuning & \multirow{5}{*}{100}& 0 & 28.78 \\
                              & EDA && 200 & 62.19 \\
                              & AugGPT && 200 & 63.42\\
                              &  More Data && 198 & 37.11 \\
                              & \graycell{\OURS} &\graycell{} & \graycell{198} & \graycell{\textbf{64.50}} \\
\midrule
\multirow{5}{*}{SNIPS}  &   Fine-tuning  & \multirow{5}{*}{70}&  0 & 60.14 \\
                              & EDA && 70 & 91.43 \\
                              & AugGPT && 70 & 89.86\\
                              &  More Data && 70 & 89.00 \\
                              & \graycell{\OURS} &\graycell & \graycell{38} & \graycell{\textbf{92.14}} \\
\midrule
\multirow{5}{*}{TREC}  &   Fine-tuning & \multirow{5}{*}{60} & 0 & 26.20 \\
                              & EDA && 120 & 72.40 \\
                              & AugGPT && 120 & 32.80\\
                              &  More Data && 138 & \textbf{89.20} \\
                              & \graycell{\OURS} &\graycell& \graycell{135} & \graycell{78.80} \\
\midrule
\multirow{5}{*}{SST-2}  &   Fine-tuning & \multirow{5}{*}{20} & 0 & 52.87 \\
                              & EDA && 40 & 63.19 \\
                              & AugGPT && 40 & 88.07\\
                              &  More Data && 40 & 72.94 \\
                              & \graycell{\OURS} &\graycell & \graycell{44} & \graycell{\textbf{92.66}} \\
\bottomrule
\end{tabular}
}
}\vspace{-2mm}
\caption{Results of \OURS compared to other baseline methods. 
Column \textit{Technique} refers to the augmentation method used as described in \sref{sec:baselines}.
Column \textit{\# Seed} indicates the size of the initial seed dataset.
Column \textit{Total Aug.} represents the total amount of \OURS data generated.
For GSM8K and CaseHOLD, we randomly sample 100 data points while for SNIPS, TREC, and SST-2, we sample 10 samples per class.
Column \textit{Acc.} indicates the final test accuracy. 
Clearly, \OURS outperforms all of synthetic baselines; even sometimes when adding in more real data from the dataset.}
\label{table:baseline}
\vspace{-4mm}
\end{table}
%%%%%%%%%%%%%%%%%%%%%%%%%%%%%%%%%%%%%%%%%%%%%%%%%%%%%%%%%%%%%%%%%%%%%%

% -----------------------------------------------------------------
\subsection{Comparison with Other Augmentation Methods}
\label{sec:baseline_results}
In \tref{table:baseline}, we compare our method against other augmentation techniques, including EDA~\cite{wei2019eda} and AugGPT~\cite{dai2023auggpt}. We also compare against adding more data from the unseen training set.
The details of all augmentation methods we used in our comparison are provided in~\sref{sec:baselines}.

On GSM8K, \OURS outperforms naive fine-tuning by over 20\%, EDA by over 8\%, and AugGPT by over 5\%.
Similarly, on CaseHOLD, \OURS outperforms the fine-tuning baseline by approximately 35\%, EDA by 2.3\%, and AugGPT by 1.1\%. 
 These improvements, particularly in comparison to AugGPT, can be attributed to \OURS's capability to generate more targeted examples based on where the model struggles, 
 as opposed to AugGPT which augments data indiscriminately. 
 This allows for more effective and targeted use of the augmented data budget.

% ------------------------------------------------------------
\subsection{Choice of Teacher Model}
\label{sec:teacher_ablation}

Thus far, we have illustrated \OURS's performance with GPT-3.5 as the teacher model, but other LLMs can serve this role as well.
A stronger teacher model is  expected to yield higher-quality augmentation and, consequently, higher accuracy. 
\tref{table:model-ablation} demonstrates the \OURS's accuracy with GPT-4-Turbo, Llama-2-70B, and Airoboros-L2-70B as the teacher model on GSM8K.
With 74 seed data examples, \OURS only achieves 11.8\% accuracy with Llama-2-70B,
which can be contrasted with 15.0\% with Airoboros and 19.8\% with GPT-4-Turbo.
This aligns with our expectation, as GPT-4-Turbo's mathematical reasoning is known to be better than
the other models, being generally on par with that of GPT-4~\cite{fu2023chain,gpt4turbovsgpt4}. 
The qualitative analysis of augmented data using different models (\fref{fig:model_new_training_data}) further supports this, 
showing that Llama and Airoboros models produce less varied data than GPT-3.5 or GPT-4-Turbo.

% ----------------------------------------------------------------
\subsection{Ablation Studies}
\label{sec:ablations}
Here, we provide ablation studies to justify the design decisions we made in \OURS.

%%%%%%%%%%%%%%%%%%%%%%%%%%%%%%%%%%%%%%%%%%%%%%%%%%%%%%%%%%%%%%%%%%%%%%%%%%%%%
\begin{table}[t!]
\centering
\footnotesize{
\setlength{\tabcolsep}{4pt}{
\begin{tabular}{c|c|c|c}
\toprule
\textbf{Dataset} & \textbf{Steps} & \textbf{Total Aug.} & \textbf{Acc. (\%)} \\
\midrule
\multirow{2}{*}{GSM8K} &
1 (one-shot) & 490 & 16.30 \\
         &  \graycell{10} (iterative) & \graycell{471} & \graycell{\textbf{23.73}} \\
\midrule
\multirow{2}{*}{CaseHOLD}  & 1 (one-shot) & 276 & 59.94 \\
         &  \graycell{10} (iterative) & \graycell{198} & \graycell{\textbf{64.50}} \\
\bottomrule
\end{tabular}
}
}
\vspace{-2mm}
\caption{Ablation on the iterative nature of \OURS with 100 seed data points. \textit{Steps} refers to the total number of augmentation steps in \OURS. For the case of 1 iteration, we prompt the teacher model to generate more samples
all at once, whereas in the 10 steps case the teacher model only generates 1 new data point per wrong example.
The results clearly show that the latter iterative approach results in better performance.}
\label{table:iterative-ablation}
\vspace{-4mm}
\end{table}

%%%%%%%%%%%%%%%%%%%%%%%%%%%%%%%%%%%%%%%%%%%%%%%%%%%%%%%%%%%%%%%%%%%%%%%%%%%%%

% ---------------------------------------------------------------------
\subsubsection{Iterative Augmentation vs One-Shot Augmentation}
\label{sec:iteration-ablation}
We first evaluate the efficacy of iterative augmentation versus adding all augmented data at once.
To evaluate this, we compare the final accuracy achieved by augmenting data over 10 iterations against adding the equivalent amount of data in a single iteration, for both the GSM8K and CaseHOLD datasets. 
As shown in~\tref{table:iterative-ablation}, using a single augmentation step with a larger amount of augmented data significantly underperforms the alternative 
of executing 10 iterative steps of \OURS with a smaller number of augmentations per iteration.
In particular, on GSM8K, augmenting one data point per example over 10 steps yields a 7.4\% higher accuracy than augmenting five data points per example in a single step.
Similarly, on CaseHOLD, iterative augmentation of one data points per example over 10 steps results in a 4.6\% improvement over a one-shot augmentation with four data points per example.
This justifies the \OURS's iterative augmentation approach that generates one data point per each incorrectly answered example.

\begin{table}[t!]
\centering
\footnotesize{
\setlength{\tabcolsep}{4.4pt}{
\begin{tabular}{c|c|c|c}
\toprule
\textbf{Dataset}  & \makecell{\textbf{Only Aug.}\\\textbf{Seed Data}} & \makecell{\textbf{Total}\\\textbf{Aug.}} & \textbf{Acc. (\%)} \\
\midrule
\multirow{2}{*}{GSM8K} & \xmark  & 4302 & 18.32 \\
                       &    \graycell\cmark & \graycell{471} & \graycell{\textbf{23.75}}\\
\midrule
    \multirow{2}{*}{CaseHOLD} &    \xmark   & 351 & 63.75 \\
                          &   \graycell \cmark& \graycell{198} & \graycell{\textbf{64.50}} \\
\bottomrule
\end{tabular}
}
}
\vspace{-2mm}
\caption{
Ablation study on whether to augment previously generated \OURS data. \textit{Only Aug. Seed Data} refers to augmenting only the seed data vs. also re-augmenting the augmented data. \textit{Total Aug.} refers to the total number of augmentations generated over 10 steps of \OURS.
}
\vspace{-4mm}
\label{table:aug_seed_data}
\end{table}
%%%%%%%%%%%%%%%%%%%%%%%%%%%%%%%%%%%%%%%%%%%%%%%%%%%%%%%%%%%%%%%%%%%%%%%%%%%%%

% ---------------------------------------------------------------------
\subsubsection{Data Augmentation with Seed Data vs Augmented Data}
\label{sec:adaptation_ablation}
In each iteration, \OURS evaluates the student model's performance only on the original seed dataset and generates augmented data from incorrect seed examples. 
However, a possible alternative is performing evaluation and data augmentation using both seed and previously augmented data.
The latter often leads to sub-optimal performance as well as excessive amounts of total augmented data points, 
as we demonstrate in~\tref{table:aug_seed_data}.
On GSM8K, generating augmented data from the previous iteration's augmented data yields 18.3\% accuracy, 
while using the seed data for further augmentation improves the accuracy to 23.75\%. 
We observe a similar trend for CaseHOLD.
As discussed in \sref{sec:llm2llm}, a potential reason for the performance drop, when using augmented data for further augmentation,
has to do with a deviation from the original data distribution.

% -----------------------------------------------
\subsubsection{From-scratch Fine-tuning vs Continuous Fine-tuning}
\label{sec:scratch_vs_continuous_ft}
Another key decision for \OURS is whether to continue fine-tuning from the last iteration's checkpoint (i.e. continuous fine-tuning) or to restart fine-tuning from the pre-trained model at each iteration (i.e. from-scratch fine-tuning). 
Considering the non-convex nature of the optimization target and complex loss landscapes, this decision is not necessarily obvious. 
Nevertheless, as shown in~\tref{table:finetune_base_model}, we observe that from-scratch fine-tuning consistently and significantly outperforms continuous fine-tuning, with up to 9\% accuracy improvement.
The inferior performance of continuous fine-tuning can be attributed to a potential overfitting to small seed data over multiple iterations of fine-tuning, especially in lower-data regimes where the seed data is small.
This can be alleviated by restarting fine-tuning from scratch in each iteration with sufficient augmented data appended to the seed data to form the training dataset.

\begin{table}[t!]
\centering
\footnotesize{
\setlength{\tabcolsep}{4.4pt}{
\begin{tabular}{c|c|c|c}
\toprule
\textbf{Dataset}  & \makecell{\textbf{From-scratch}\\\textbf{Fine-tuning}} & \makecell{\textbf{Total}\\\textbf{Aug.}} & \textbf{Acc. (\%)} \\
\midrule
\multirow{2}{*}{GSM8K} 
                       & \xmark & 230 & 14.71 \\
                       & \graycell \cmark & \graycell{471} & \graycell{\textbf{23.75}}\\
\midrule
    \multirow{2}{*}{CaseHOLD} 
                        & \xmark & 154 & 60.50 \\
                        & \graycell \cmark & \graycell{198} & \graycell{\textbf{64.50}} \\
\bottomrule
\end{tabular}
}
}
\vspace{-2mm}
\caption{
Ablation study on whether to fine-tune from scratch or to do continuous fine-tuning. \textit{From-scratch Fine-tuning} refers to whether we fine-tune the base model from scratch vs. fine-tune the previous step's model. \textit{Total Aug.} refers to the total number of augmentated examples generated over 10 steps of \OURS.}
\vspace{-4mm}
\label{table:finetune_base_model}
\end{table}
%%%%%%%%%%%%%%%%%%%%%%%%%%%%%%%%%%%%%%%%%%%%%%%%%%%%%%%%%%%%%%%%%%%%%%%%%%%%%

\section{Conclusion}

We have introduced \OURS, an adaptive and iterative LLM-based data augmentation framework that uses LLMs to scale up smaller fine-tuning datasets in lieu of manually generating more data.
This framework substantially reduces the amount of real data needed, and it allows us to efficiently scale the dataset with synthetic data that can match or even exceed the effect of hand-collecting more data.
The method is effective because of the iterative and targeted nature of the process, which allows us to boost the signal from data points that the LLM gets wrong.
As a result, we were able to achieve a 24.2\% improvement on GSM8K, 32.6\% on CaseHOLD, 32.0\% on SNIPS, 52.6\% on TREC, and 39.8\% on the SST-2 dataset in the low-data regime using a Llama-2-7B student model.
Future work can focus on tuning the hyperparameters of our framework as well as incorporating our approach with other LLM techniques such as prompt tuning and few-shot learning.

\section*{Limitations}

Our results primarily reflect improvements that occur in a low training data regime, from tens of examples to a couple thousand. 
However, practitioners may deal with larger datasets from time to time, in which our method may be out of scope.

Furthermore, there could be other factors that help explain the disparity in performance between different teacher models. 
Also, we have analyzed the generated data for differences in quality, but there may be other ways to close the gap between open-source models and the GPT models as a teacher model.
This warrants further investigation.

Our focus primarily reflects a specific use case where there is low training data available due to difficulty in data collection such as labor or resource constraints. 
Exploring the effects of using synthetic data to further eke out performance when there is abundant data is a promising research direction.

\section*{Ethics Statement}

\OURS relies on using LLMs to augment a training dataset in order to train another student LLM more efficiently. This can reduce the energy and monetary cost of experimentation and machine learning research, as it enables those with smaller datasets to achieve better performance on a domain-specific task. Of course, misuse of this method may lead to unethical data being generated, which can lead to societal harm. This is not a concern specific to this work, but to LLM research in general. Furthermore, there are still open questions about latent implicit biases and ethical issues surrounding the generated output of LLMs that the authors and practitioners of this method are aware of and continue to consider throughout the whole process.

\section*{Acknowledgements}
We appreciate the valuable feedback from Andrew Aikawa. 
We acknowledge gracious support from Furiosa team.
We also appreciate the support from Microsoft through their Accelerating Foundation Model Research, including
great support from Sean Kuno.
Furthermore, we appreciate support from
Google Cloud, the Google TRC team, and specifically Jonathan Caton, and Prof. David Patterson.
Prof. Keutzer's lab is sponsored by the Intel corporation, Intel One-API, Intel VLAB team, the Intel One-API center of
excellence, Apple, Samsung, Panasonic, as well as funding through BDD and BAIR.
We appreciate great feedback and support from Ellick Chan, Saurabh Tangri, Andres
Rodriguez, and Kittur Ganesh.
Sehoon Kim  would like to acknowledge the support from the Korea Foundation for Advanced Studies (KFAS).
Amir Gholami was supported through funding from Samsung SAIT.
Michael W. Mahoney would also like to acknowledge
a J. P. Morgan Chase Faculty Research Award 
as well as 
the DOE, NSF, and ONR.
Our conclusions do not necessarily reflect the position or the policy of our sponsors, and no official endorsement should be~inferred.

% % Entries for the entire Anthology, followed by custom entries
\bibliography{anthology,custom}
\bibliographystyle{acl_natbib}
\clearpage

% These commands are here for proper references to material in the appendix
\renewcommand\thefigure{\thesection.\arabic{figure}} % Redefine figure numbering
\setcounter{figure}{0} % Reset figure counter

\renewcommand\thetable{\thesection.\arabic{table}} % Redefine table numbering
\setcounter{table}{0} % Reset table counter
\appendix

\section{Experimental Setup}
\label{sec:experimental_setup_appendix}

% -------------------------------------------------------
\subsection{Datasets}

We evaluate \OURS on five different datasets that are either multiple-choice or classification tasks and were widely adopted in prior works including \cite{ubani2023zeroshotdataaug}. 
Our datasets are as follows:

\begin{enumerate} [leftmargin=3mm]
\vspace{-1mm}
    \item \textbf{GSM8K}: a grade school math work multiple-choice dataset that consists of 7.5K train problems and 1K test problems~\cite{cobbe2021gsm8k}.
    \fref{fig:gsm8k_example} shows an example.
    \vspace{-1mm}
    \item \textbf{CaseHOLD}: a multiple-choice law dataset that requires one to choose the relevant holding (i.e. the court's determination) of a cited case which backs up the proceeding argument~\cite{zhengguha2021}.
    \fref{fig:casehold_example} shows an example.
    \item \textbf{SNIPS}: a 7-way classification dataset to determine the correct user intent for a voice assistant.~\cite{coucke2018snips}. \fref{fig:snips_example} shows an example.
    \item \textbf{TREC}: a 6-way classification dataset where one must classify the type of text into a category e.g., abbreviation, location, or numeric value~\cite{li-roth-2002-learning}. \fref{fig:trec_example} shows an example.
    \item \textbf{SST-2}: a binary classification dataset to decide whether a sentence of positive or negative sentiment~\cite{socher2013recursive}. \fref{fig:sst2_example} shows an example.
\end{enumerate}

For each dataset, we sample between 0.02\% to 50\%
of the total training data and use this as the seed data for each experiment.
This allows us to measure how effectively \OURS scales up small task-specific datasets.
For consistency, we use identical samples of seed data across different experiments (e.g. 1\% on GSM8K) to avoid introducing new randomness with different samples.

In particular, for SNIPS, TREC, and SST-2, we always uniformly sample the same number of examples per class, similar to \cite{dai2023auggpt, ubani2023zeroshotdataaug}.
In \tref{table:results}, we sample 10, 15, and 20 examples per class to measure the efficacy of \OURS in the extreme low-data regime.
These three tasks are relatively simpler than GSM8K and CaseHOLD, and therefore using an extremely small amount of training data is sufficient to achieve exemplary performance.

% -------------------------------------------------------
\subsection{Models}
For all of our experiments, we use Llama-2-7B \cite{touvron2023llama2} as the student model. 
We perform minimal prompt tuning for each task, only formatting the data as necessary and not employing many few-shot examples, which can be seen in \fref{fig:gsm8k_example} and  \fref{fig:casehold_example}.
Using excessive prompting would undermine the benefits of fine-tuning and would muddle the evaluation of the effectiveness of \OURS.
Our fine-tuning settings are described in \sref{sec:finetuning_settings}.

For our main experiments, we use GPT-3.5 (1106 release) as the teacher model for data generation.
In \sref{sec:teacher_ablation} and \tref{table:model-ablation}, we show that our framework can be extended to different teacher models such as the more powerful GPT-4-Turbo (1106 release) model as well as open source LLMs such as Llama-2-70B-chat~\cite{touvron2023llama2} and Airoboros-l2-70b-3.1.2~\cite{airoboros}.

% -------------------------------------------------------
\subsection{Baselines and Evaluation}
\label{sec:baselines}
To measure the efficacy of \OURS, we fine-tune the student model using samples of different sizes from each dataset and evaluate on the validation sets of each of these datasets.
We then run 10 steps of \OURS, and use the validation sets to select the best performing model.
In \sref{sec:main_results}, we compare these results against basic fine-tuning on just the seed data.

For GSM8K and TREC, since there is no development set, we choose the best checkpoint's test set results to be representative for the overall improvement.
Similarly, for SST-2, since the test set labels are not public, we use the development set results.
For all other datasets, we record the test set performance of the checkpoint that performs best on the development set.

For TREC, SST-2, and SNIPS, since these are simple classification tasks, we perform string matching between the generated output and the ground truth after some cleanup. For CaseHOLD, which is a multiple choice task, we extract the letter of the answer that the model generates. For GSM8K, we use a regular expression extraction based on the answer format that GSM8K provides. Specifically, we extract the number after the \#\#\#\# token.

In \sref{sec:baseline_results}, we only sample 100 examples from GSM8K and CaseHOLD.
For SNIPS, TREC, and SST-2, we sample 10 examples per class.
We run an extensive ablation study against several different baselines:
\begin{itemize}[leftmargin=3mm]
\vspace{-1mm}
    \item \textbf{Fine-tuning}: Standard fine-tuning on the initial seed data.
    \vspace{-1mm}
\item \textbf{EDA}: EDA~\cite{wei2019eda} which uses synonym replacement plus random swap, insert, and deletion to augment text with $10\%$ probability.
Note that we take care not to augment any special formatting or structural elements for each dataset.
    \vspace{-1mm}
\item \textbf{AugGPT}: Augment our seed data using the teacher prompts from \sref{sec:llm2llm} with no filtering, similar to \cite{dai2023auggpt, ubani2023zeroshotdataaug, yoo2021gpt3mix}.
    \vspace{-1mm}
\item \textbf{More Data}: Randomly sampling new data from unseen train data and adding to the training data.
\end{itemize}

\subsection{Fine-tuning Settings}
\label{sec:finetuning_settings}
Following the example from Alpaca \cite{taori2023alpaca}, all models are fine-tuned with a maximum sequence length of $1024$ for $3$ epochs with a batch size of $128$ examples. We use a learning rate of $2\times10^{-5}$ with $0$ weight decay and a warmup ratio of $0.03$ with a cosine learning rate scheduler. 
These models were trained using either 4 NVIDIA A100-80GB or 8 NVIDIA A6000s.
We do full fine-tuning for simplicity and to reduce the complexity of our experiments, but in practice, one can also use some form of parameter-efficient fine-tuning method such as LoRA \cite{hu2021lora}.

\begin{table*}[t!]
\centering
\footnotesize{
\setlength{\tabcolsep}{6pt}{
\begin{tabular}{c|c|c|c|c}
\toprule
{\textbf{Dataset}} & \textbf{\# Seed} & \textbf{Teacher} & \textbf{Total \# Aug} & \textbf{Accuracy (\%)} \\
\midrule
\multirow{12}{*}{GSM8K} & \multirow{4}{*}{74 (1\%)} & Llama-2-70B & 333 & 11.83 \\
                       &       & Airoboros & 345 & 15.01 \\
                       &       & GPT-3.5 & 391 & 19.56\\
                       &       & GPT-4-Turbo & 388 & \textbf{19.79} \\
\cmidrule{2-5}
& \multirow{4}{*}{149 (2\%)} 
                              & Llama-2-70B & 661 & 17.59 \\
                       &       & Airoboros & 671 & 19.33 \\
                       & & GPT-3.5 & 802 & 25.70 \\
                       &       & GPT-4-Turbo & 805 & \textbf{25.78} \\
\cmidrule{2-5}
& \multirow{4}{*}{343 (5\%)}  
                              & Llama-2-70B & 1308 & 19.33 \\
                       &       & Airoboros & 1286 & 21.76\\
                       & & GPT-3.5 & 1641 & 27.07 \\
                       &       & GPT-4-Turbo & 1739 & \textbf{28.43} \\
\bottomrule
\end{tabular}
}
}
\caption{Experiments on how the quality of teacher model affects the performance of \OURS. For each of these experiments, we only change the teacher model to measure the effect of the teacher model on the final outcome.}
\label{table:model-ablation}
\end{table*}

\section{\OURS Details}

\subsection{Fine-tuning}
\label{sec:finetuning}
The fine-tuning step trains a small student model using seed data and previously generated \OURS data, if any.
This \OURS data is generated by a process further detailed in \sref{sec:llm2llm} that generates synthetic data targeted toward data points the model got wrong.
We always fine-tune the original student model on the full dataset (seed data + \OURS data) at each step. 
Our ablation study in \sref{sec:scratch_vs_continuous_ft} shows that fine-tuning the original baseline model from scratch on the full dataset always outperformed re-using the already-fine-tuned model.
We hypothesize that this is because fine-tuned models have already seen most of the data, causing them to overfit and fail to converge to a better optimum.

\subsection{Evaluation}
\label{sec:evaluation}
After fine-tuning, the model needs to be evaluated on the original (training) seed data
to identify the examples that the model gets wrong.
This allows the \OURS framework to use those failed examples to generate targeted synthetic data for the model to train on. 
For example, if the model was unable to solve problems involving the Pythagorean theorem as in \fref{fig:autollm}, these examples will be used to generate more problems with this concept.

For many datasets, this evaluation step can be extremely costly, as traditional NLP datasets can have more than thousands of data points.
However, this evaluation step is cost-effective and relatively quick in the low-data regime where the seed dataset size is small, thereby not slowing down the \OURS process.

\subsection{Filtering Generated Dataset}
\label{sec:filtering}

Once the teacher model generates the synthetic data, we need to apply simple filtering of the output for quality insurance.
Like in previous work \cite{wang-etal-2023-self-instruct, taori2023alpaca}, we use regex filters to ensure that the basic format of the output is aligned with our expectations. 
We also use a ROUGE \cite{lin2004rouge} filter in order to enforce that the augmented data points are sufficiently different from previous samples.
However, we use a weaker ROUGE filter of $0.95$ to filter out similar instructions, rather than the score of $0.85$ used in other works like Alpaca \cite{taori2023alpaca} and Self-instruct \cite{wang-etal-2023-self-instruct}. 
The reason we can do this is because unlike Alpaca and Self-Instruct, we do not require as much diversity in the generations, as we are not targeting general-purpose instruction-following.
In fact, we would like to constrain the generated data points to the task and domain of the datasets as much as we can.
Thus, we are able to use a weaker filter so that we can simply filter out exact matches during generation.
In addition, since we are augmenting each sample individually, there is already enough inherent diversity in the generation process.

\subsection{Prompting Details}
\label{sec:teacher_prompting}
We devised simple but thorough prompts for each task that the teacher model uses while augmenting the dataset.
Previous work in open-domain dataset generation such as \cite{wang-etal-2023-self-instruct, mukherjee2023orca, xu2023wizardlm} used generic system instructions for generating new data points from stronger models such as GPT-3.5 and GPT-4. 
This was necessary as these approaches targeted improving the LLM over a wide range of different tasks.
However, for \OURS, we are trying to improve the LLM at domain-specific tasks.
Thus, for each task, the system prompt that we give to the teacher-model differs on a per-task basis.
This allows the user to inject and leverage domain-specific knowledge about the task into the dataset generation procedure, creating higher quality fine-tuning data.
In practice, we also use in-context learning with few-shot prompting to bootstrap the teacher model's ability to generate relevant questions.

The detailed system prompt and in-context examples for each dataset are provided below:

\begin{enumerate}[leftmargin=5mm]
\vspace{-1mm}
    \item GSM8K: System~(\fref{fig:gsm8k_system_prompt}) and In-Context Examples~(\fref{fig:gsm8k_user_prompt})
    \vspace{-1mm}
    \item CaseHOLD: System~(\fref{fig:casehold_system_prompt}) and In-Context Examples~(\fref{fig:casehold_user_prompt})
    \vspace{-1mm}
    \item SNIPS: System~(\fref{fig:snips_system_prompt}) and In-Context Examples~(\fref{fig:snips_user_prompt})
    \vspace{-1mm}
    \item TREC: System~(\fref{fig:trec_system_prompt}) and In-Context Examples~(\fref{fig:trec_user_prompt})
    \vspace{-1mm}
    \item SST-2: System~(\fref{fig:sst2_system_prompt}) and In-Context Examples~(\fref{fig:sst2_user_prompt})
\end{enumerate}

\subsection{Training and Data Generation Costs}
\label{sec:training_costs} 

In \tref{table:training_costs}, we report the training and data generation costs to perform \OURS. 
This includes the cost of generating new data from OpenAI as well as the amount of GPU hours required to train and evaluate the student models.
We measured these numbers using 4xA100-80GB PCIe based NVIDIA GPUs. 
As we can see, generating data for \OURS costs relatively little compared to the cost of collecting new data points manually.
Furthermore, the process of fine-tuning the student model also finishes in a reasonable amount of time. 

\begin{table}[t]
\centering
\footnotesize{
\setlength{\tabcolsep}{3.5pt}{
\begin{tabular}{c|c|c|c}
\toprule
\textbf{Dataset} &  \textbf{\% Data}  & \textbf{Cost (\$)} & \textbf{Time (Hours)} \\
\midrule
\multirow{3}{*}{GSM8K} &  1\% & 0.35 & 3.28 \\
                        & 5\% & 1.48 & 9.07 \\
                        & 10\% & 3.64 & 14.54 \\
\midrule
\multirow{3}{*}{CaseHOLD} &1\% &	1.50 &	6.68 \\
&5\%	&0.84	&16.87\\
&10\%	&2.19	&31.95\\
\midrule
\multirow{3}{*}{SNIPS} &0.5\%&	0.02&	0.85 \\
&0.8\%&	0.05&	1.29 \\
&1\%	&0.05	&1.40 \\

\midrule
\multirow{3}{*}{TREC} &1.1\%&	0.05&	0.67 \\
&1.6\%&	0.01&	0.44 \\
&2.2\%&	0.02&	0.61 \\
\midrule
\multirow{3}{*}{SST-2}	&0.02\%	&0.01	&0.54 \\
&0.04\%	&0.01	&0.80 \\
&0.06\%	&0.00	&0.64 \\

\bottomrule
\end{tabular}
}
}\vspace{-2mm}
\caption{Training and data generation Costs of \OURS. The first and second columns indicate the dataset and percentage of the training data used as initial seed data for that experiment. The third column indicates the total cost to generate the data from the GPT-3.5 teacher model. The fourth column shows the total time in hours to train and evaluate the student model. As we can see, data generation costs for \OURS are relatively little compared to the cost of manually curating new data. Furthermore, fine-tuning and evaluation of the student model finishes in a reasonable time.}
\label{table:training_costs}
\vspace{-4mm}
\end{table}

\subsection{Decontamination Experiments}
\label{sec:word_overlap} 

%%%%%%%%%%%%%%%%%%%%%%%%%%%%%%%%%%%%%%%%%%%%%%%%%%%%%%%%%%%%%%%%%%%%%%
\begin{table}[t!]
\centering
\footnotesize{
\setlength{\tabcolsep}{6pt}{
\begin{tabular}{c|c|c|c}
\toprule
Dataset  & Avg. & >66\% (\%) & Max \% (Count)  \\
\midrule
    GSM8K Train     &  52.08    &   11.46    &   81.48 (1)    \\
    GSM8K Test      &  26.38    &       0    &   46.67 (1)  \\
\midrule
    CaseHOLD Train  &  31.01    &    0     &   50.42 (2)     \\
    CaseHOLD Test   & 25.66     &     0    &   38.46 (1)  \\
\midrule
    SNIPS Train     &  59.94    &   34.21   &   85.71 (1)       \\
    SNIPS Test      & 45.90     &  7.89 &   80.00  (1)  \\
\midrule
    TREC Train      &  47.62    &   17.04   &   80.00 (4)        \\
    TREC Test       &  33.47    &  6.67    &   80.00 (1)  \\
\midrule
    SST-2 Train     &  34.35    &  0        &   57.14 (2)      \\
    SST-2 Dev      & 25.46     &  0        &  42.86 (1)  \\
\bottomrule
\end{tabular}
}
}
\caption{Word overlap by dataset of synthetic examples generated after 10 steps of \OURS using 100 seed examples as in \sref{sec:baseline_results}. 
Column \textit{Dataset} indicates the dataset and train split being used to compare the synthetic data with.
Column \textit{Avg.} is the average overlap percentage. Column \textit{>66\%} is the percentage of examples with above 66\% overlap, and column \textit{Max \% (Count)} indicates the maximum overlap percentage and the number of examples at that overlap percentage.}
\label{table:word_overlap}
\end{table}
%%%%%%%%%%%%%%%%%%%%%%%%%%%%%%%%%%%%%%%%%%%%%%%%%%%%%%%%%%%%%%%%%%%%%%

When using an LLM to generate data, there are potential concerns of data contamination i.e., when the teacher LLM has been trained on the test data and thus leaks it into the student model's training data which would artificially inflate the student model's scores.
To test for this we measure the word overlap percentage between examples generated by the teacher model and examples from the train and test set for each dataset in the same manner as~\cite{ubani2023zeroshotdataaug}.  

Word overlap is computed by first removing stop words and punctuation from each example. Then, for each example pair, we count the number of words that are common between the two, and divide by the number of words in longer example. This provides a metric for measuring how many words are similar between the two examples.

For each dataset, we run \OURS for 10 steps using the same amount of seed data as in \sref{sec:baseline_results}. Then, we take all the synthetic data generated for the final step and calculate our word overlap metric per example. We then calculate word overlap summary statistics similarly to~\cite{ubani2023zeroshotdataaug} which are the percentage of examples above 66\% similarity, the maximum overlap, and the number of examples at the maximum overlap.

Looking at \tref{table:word_overlap}, we can see varying levels of overlap between the synthetic data and the training data.
This is to be expected, as \OURS uses training data as seed data in order to generate new synthetic examples based on that training data.

Regarding the test/dev set results, we see that for GSM8K, CaseHOLD, and SST-2, 0\% of the generated examples have above 66\% word overlap with the test set. This indicates that the model is not leaking test set data into the student model's training data. For SNIPS and TREC, we see that the percent of examples with above 66\% overlap is still well below 10\%  thus indicating no large-scale overlap.

For SNIPS and TREC specifically the maximum overlap percentage is 80\%. Upon closer inspection this is because these two datasets have very short examples. Given the small number of such examples we don't believe this poses a high risk of data contamination. \cite{ubani2023zeroshotdataaug} came to a similar conclusion regarding these two datasets.

%%%%%%%%%%%%%%%%%%%%%%%%%%%%%%%%%%%%%%%%%%%%%%%%%%%%%%%%%%%%%%%%%%%%%%

%%%%%%%%%%%%%%%%%%%%%%%%%%%%%%%%%%%%%%%%%%%%%%%%%%%%%%%%%%%%%%%%%%%%%%

\begin{figure*}[t!]
\centering
\begin{tcolorbox}[colback=gray!5!white,colframe=gray!75!black]
\textbf{GSM8K Example:} 
\\~\\
Victor, Austin, and Brian made traps to catch shrimp. Victor's trap caught 26 shrimp and Austin's trap caught 8 less than Victor's. Brian's trap caught half of Victor and Austin's total number of shrimp. If the boys then sold their shrimp for \$7 for every 11 tails of shrimp and then divided their earnings equally amongst themselves, how much money does each boy make?
\\~\\
Austin's trap caught 26 - 8 = <<26-8=18>>18 shrimp.\\Together, Victor and Austin's traps caught 18 + 26 = <<18+26=44>>44 shrimp.\\Brian's trap caught 44/2 = <<44/2=22>>22 shrimp\\In total, they caught 26 + 18 + 22 = <<26+18+22=66>>66 shrimp.\\They were able to sell 66/11 = <<66/11=6>>6 sets of shrimp.\\They made a total of 6 x $7 = $<<6*7=42>>42\\Each boy made $42/ 3 = $<<42/3=14>>14\\\#\#\#\# 14
\end{tcolorbox}
    \caption{Formatted example from GSM8K.}
    \label{fig:gsm8k_example}
\end{figure*}
%%%%%%%%%%%%%%%%%%%%%%%%%%%%%%%%%%%%%%%%%%%%%%%%%%%%%%%%%%%%%%%%%%%%%%

%%%%%%%%%%%%%%%%%%%%%%%%%%%%%%%%%%%%%%%%%%%%%%%%%%%%%%%%%%%%%%%%%%%%%%
\begin{figure*}[t!]
    \centering
\begin{tcolorbox}[colback=gray!5!white,colframe=gray!75!black]
\textbf{CaseHOLD Example:}
\\~\\
The following context is from a judicial decision where the holding statement has been masked out as <HOLDING>.\\    \\Context: from behind the bench in the robe is protected by the First Amendment, even if his use of the trappings of judicial office were notprotected by First Amendment); Halleck v. Berlinger, 427 F. Supp. 1225, 1241 (D.D.C. 1977) (applying the First Amendment in disciplinary proceeding to comments made from the bench, but finding the particular comments outside of its protection); Mississippi Comm’n on Judicial Performance v. Boland, 975 So.2d 882, 891-92 (Miss. 2008) (applying First Amendment to ajudge acting in her “capacity as a justice court judge” at a conference seeking certification to start a drug court, but held that First Amendment did not apply because judge’s insulting comments were not matters of “legitimate public concern.”); In re Rome, 218 Kan. 198, 542 P.2d 676, 684 (1975) (<HOLDING>). 11 As indicated in the Gentile syllabus,\\\\Please select the correct holding statement from the options below.\\\\A. holding that free speech protection of new jersey constitution requires subject to reasonable restrictions privatelyowned shopping centers to permit speech on political and societal issues on premises unlike first amendment of federal constitution\\B. recognizing that code is speech\\C. holding that first amendment protections apply to compelled speech as well as restrictions on speech\\D. holding that although ajudge has the right of free speech any restrictions placed by the code of professional responsibility are acceptable limits and prevent the first amendment from exempting a judge from discipline for proven judicial misconduct\\E. holding that the first amendment limits judicial discretion to seal documents in a civil case
\\~\\
D
\end{tcolorbox}
    \caption{Formatted example from CaseHOLD.}
    \label{fig:casehold_example}
\end{figure*}
%%%%%%%%%%%%%%%%%%%%%%%%%%%%%%%%%%%%%%%%%%%%%%%%%%%%%%%%%%%%%%%%%%%%%%

%%%%%%%%%%%%%%%%%%%%%%%%%%%%%%%%%%%%%%%%%%%%%%%%%%%%%%%%%%%%%%%%%%%%%%
\begin{figure*}[t!]
    \centering
\begin{tcolorbox}[colback=gray!5!white,colframe=gray!75!black]
\textbf{SNIPS Example:}
\\~\\
The following is a transcript of something someone said.\\Classify the intent of the speaker into the following categories:\\- AddToPlaylist\\- BookRestaurant\\- GetWeather\\- PlayMusic\\- RateBook\\- SearchCreativeWork\\- SearchScreeningEvent\\\\Transcript: go to bioruby
\\~\\
SearchCreativeWork
\end{tcolorbox}
    \caption{Formatted example from SNIPS.}
    \label{fig:snips_example}
\end{figure*}
%%%%%%%%%%%%%%%%%%%%%%%%%%%%%%%%%%%%%%%%%%%%%%%%%%%%%%%%%%%%%%%%%%%%%%

%%%%%%%%%%%%%%%%%%%%%%%%%%%%%%%%%%%%%%%%%%%%%%%%%%%%%%%%%%%%%%%%%%%%%%
\begin{figure*}[t!]
    \centering
\begin{tcolorbox}[colback=gray!5!white,colframe=gray!75!black]
\textbf{TREC Example:}
\\~\\
The following is a question.\\Classify the question into the following categories:\\- ABBR\\- ENTY\\- DESC\\- HUM\\- LOC\\- NUM\\\\Question: What is the full form of .com ?
\\~\\
ABBR
\end{tcolorbox}
    \caption{Example from TREC.}
    \label{fig:trec_example}
\end{figure*}
%%%%%%%%%%%%%%%%%%%%%%%%%%%%%%%%%%%%%%%%%%%%%%%%%%%%%%%%%%%%%%%%%%%%%%

%%%%%%%%%%%%%%%%%%%%%%%%%%%%%%%%%%%%%%%%%%%%%%%%%%%%%%%%%%%%%%%%%%%%%%
\begin{figure*}[t!]
    \centering
\begin{tcolorbox}[colback=gray!5!white,colframe=gray!75!black]
\textbf{SST-2 Example:}
\\~\\
Classify the following movie review as positive or negative: dazzling and sugar-sweet , a blast of shallow magnificence that only sex , scandal , and a chorus line of dangerous damsels can deliver .
\\~\\
negative
\end{tcolorbox}
    \caption{Example from SST-2.}
    \label{fig:sst2_example}
\end{figure*}
%%%%%%%%%%%%%%%%%%%%%%%%%%%%%%%%%%%%%%%%%%%%%%%%%%%%%%%%%%%%%%%%%%%%%%

%%%%%%%%%%%%%%%%%%%%%%%%%%%%%%%%%%%%%%%%%%%%%%%%%%%%%%%%%%%%%%%%%%%%%%
\begin{figure*}[t!] 
\centering
\begin{tcolorbox}[colback=gray!5!white,colframe=gray!75!black]
\textbf{System:}
\\~\\
You are a educational A.I. whose purpose is to take math problems that students get wrong and generate new problems to help them practice their mathematical skills. Your goal is to generate a set of new math problems that reflect the different skills and techniques found in the example problem.\\\\Here are the requirements:\\1. A GPT language model should be able to complete the problem. For example, do not ask the assistant to create any visual or audio output. For another example, do not ask the assistant to wake you up at 5pm or set a reminder because it cannot perform any action.\\2. The math problem should be in English.\\3. The output should be an appropriate response to the question. Make sure the output is less than 100 words.\\4. The answer to the problem should be expressed as a number, not a fraction. For example, if the answer is one-half, return 0.5, not 1/2 or "one half".\\5. The answer to the problem should not have units i.e. if the answer is 6 cups, just write 6 as the [ANSWER]\\6. Always include some calculation to show your work for how you got your ANSWER.\\7. Don't make any mathematical mistakes of your own!\\8. Try not to copy too much information from the original problem. If you must, try and replace names and numbers so that we can test the student's understanding, rather than their ability to memorize previous test questions.\\\\Always return your instructions in the form:\\1. Question: [QUESTION]\\Answer: [CALCULATION]\\\#\#\#\# [ANSWER]
\end{tcolorbox}
    \caption{System Prompt for GSM8K Generation}
    \label{fig:gsm8k_system_prompt}
\end{figure*}
%%%%%%%%%%%%%%%%%%%%%%%%%%%%%%%%%%%%%%%%%%%%%%%%%%%%%%%%%%%%%%%%%%%%%%

%%%%%%%%%%%%%%%%%%%%%%%%%%%%%%%%%%%%%%%%%%%%%%%%%%%%%%%%%%%%%%%%%%%%%%
\begin{figure*}[t!] 
    \centering
\begin{tcolorbox}[colback=gray!5!white,colframe=gray!75!black]
\textbf{User:}
\\~\\
The student was given the following question:\\\\Betty is saving money for a new wallet which costs \$100. Betty has only half of the money she needs. Her parents decided to give her \$15 for that purpose, and her grandparents twice as much as her parents. How much more money does Betty need to buy the wallet?\\\\The answer key has this as the rationale and answer:\\\\In the beginning, Betty has only 100 / 2 = \$<<100/2=50>>50.\\Betty's grandparents gave her 15 * 2 = \$<<15*2=30>>30.\\This means, Betty needs 100 - 50 - 30 - 15 = \$<<100-50-30-15=5>>5 more.\\\#\#\#\# 5\\\\Please generate 1 similar question, along with the correct calculations and rationale.
\\~\\
\textbf{Assistant:}
\\~\\
Sure, I can help you with that! Here's a new math question based on the same scenario, along with the correct calculations and rationale:\\\\1. Question: Betty is saving money for a new wallet which costs \$150. Betty has already saved \$30. Her parents decided to give her \$20 for that purpose, and her grandparents gave her three times as much as her parents. How much more money does Betty need to buy the wallet?\\Answer: In the beginning, Betty has saved \$30. Betty's parents gave her \$20. Betty's grandparents gave her 3 * \$20 = \$60. Total amount Betty has: \$30 + \$20 + \$60 = \$110. The cost of the wallet is \$150. Amount more needed: \$150 - \$110 = \$40.\\\#\#\#\# 40
\end{tcolorbox}
    \caption{In-Context Example for GSM8K Generation}
    \label{fig:gsm8k_user_prompt}
\end{figure*}
%%%%%%%%%%%%%%%%%%%%%%%%%%%%%%%%%%%%%%%%%%%%%%%%%%%%%%%%%%%%%%%%%%%%%%

%%%%%%%%%%%%%%%%%%%%%%%%%%%%%%%%%%%%%%%%%%%%%%%%%%%%%%%%%%%%%%%%%%%%%%
\begin{figure*}[t!] 
    \centering
\begin{tcolorbox}[colback=gray!5!white,colframe=gray!75!black]
\textbf{System:}
\\~\\
You are LawGPT, an AI agent who knows everything there is to know about U.S. law.\\You know the result of every court case and you know every law in the lawbook.\\The user is trying to choose the correct holding of the case given the context and argument of the court.\\You are trying to give the user assistance by giving them more practice questions for the questions that they get wrong.\\\\Here are the requirements:\\1. A GPT language model should be able to complete the problem. For example, do not ask the assistant to create any visual or audio output. For another example, do not ask the assistant to wake you up at 5pm or set a reminder because it cannot perform any action.\\2. The context, holding, and options should be in english.\\3. The questions that you generate should test for whether the user understands the case names and their holdings and whether the user can re-frame relevant holdings to backup the argument in the context.\\4. The context should always end with a citation such as "See United States v. Newman, 125 F.3d 863 (10th Cir.1997) (unpublished) (<HOLDING>); United States v. Dodge, 846 F.Supp. 181,"\\5. The citation absolutely needs to have the mask phrase <HOLDING> which is the place where the legal holding would normally be.\\6. The questions should always be multiple choice.\\7. There should always be 5 options: 1 options should be a holding that backs up the argument in the context, the other 4 should be sufficiently different. Each option has to start with the word "holding"\\8. There can only be 1 answer: A, B, C, D, or E.\\9. Don't make any mistakes matching the holdings yourself.\\10. Try not to copy too much information from the original problem. You don't want the user to just memorize their answer.\\11. Make the context similar to the context in question, make sure that the holding that is being tested is the same.\\12. The wrong answer choices can be any other reasonable holding, but it should be sufficiently different from the correct answer.\\13. Do not make your context too short. Remember, these arguments in the context are being made by judges and should look like they were written by a judge.\\\\Always return your instructions in the form:\\1. Context: [CONTEXT]\\\\Please select the correct holding statement from the options below.\\\\A. [OPTION 1]\\B. [OPTION 2]\\C. [OPTION 3]\\D. [OPTION 4]\\E. [OPTION 5]\\Answer: [ANSWER]
\end{tcolorbox}
    \caption{System Prompt for CaseHOLD Generation}
    \label{fig:casehold_system_prompt}
\end{figure*}
%%%%%%%%%%%%%%%%%%%%%%%%%%%%%%%%%%%%%%%%%%%%%%%%%%%%%%%%%%%%%%%%%%%%%%

%%%%%%%%%%%%%%%%%%%%%%%%%%%%%%%%%%%%%%%%%%%%%%%%%%%%%%%%%%%%%%%%%%%%%%
\begin{figure*}[t!] 
    \centering
\begin{tcolorbox}[colback=gray!5!white,colframe=gray!75!black]
\small
\textbf{User:}
\\~\\
The following is a multiple choice question about the holding statements of a judicial decision that the user got wrong including the correct answer from the answer sheet:\\\\Context: Drapeau’s cohorts, the cohort would be a “victim” of making the bomb. Further, firebombs are inherently dangerous. There is no peaceful purpose for making a bomb. Felony offenses that involve explosives qualify as “violent crimes” for purposes of enhancing the sentences of career offenders. See 18 U.S.C. § 924(e)(2)(B)(ii) (defining a “violent felony” as: “any crime punishable by imprisonment for a term exceeding one year ... that ... involves use of explosives”). Courts have found possession of a'bomb to be a crime of violence based on the lack of a nonviolent purpose for a bomb and the fact that, by its very nature, there is a substantial risk that the bomb would be used against the person or property of another. See United States v. Newman, 125 F.3d 863 (10th Cir.1997) (unpublished) (<HOLDING>); United States v. Dodge, 846 F.Supp. 181,\\\\Please select the correct holding statement from the options below.\\\\A. holding that possession of a pipe bomb is a crime of violence for purposes of 18 usc  3142f1\\B. holding that bank robbery by force and violence or intimidation under 18 usc  2113a is a crime of violence\\C. holding that sexual assault of a child qualified as crime of violence under 18 usc  16\\D. holding for the purposes of 18 usc  924e that being a felon in possession of a firearm is not a violent felony as defined in 18 usc  924e2b\\E. holding that a court must only look to the statutory definition not the underlying circumstances of the crime to determine whether a given offense is by its nature a crime of violence for purposes of 18 usc  16\\Answer: A\\\\Please generate 1 similar question, along with 5 different holding options and the correct answer.
\\~\\
\textbf{Assistant:}
\\~\\
Sure, I can help you with that! Here's a new case holding question based on that question:\\\\1. Context: In the case of Martinez v. State, it was determined that the act of threatening another with a weapon, even if not used, constitutes a violent act. This is based on the premise that the mere possession of a weapon with the intent to threaten establishes an inherent risk of physical harm. Such acts align with the definition of "violent crimes" which can be used as grounds for stricter penalties under the jurisdiction of certain statutes. Refer to 18 U.S.C. § 16(b) which defines a “crime of violence” as: “any other offense that is a felony and that, by its nature, involves a substantial risk that physical force against the person or property of another may be used in the course of committing the offense.” In the context of this, the Jones v. Commonwealth, 279 F.3d 722 (4th Cir.2001) case had a significant <HOLDING>.\\\\Please select the correct holding statement from the options below.\\\\A. holding that mere possession of a weapon without any intent does not constitute a crime of violence under 18 U.S.C. § 16(b)\\B. holding that using a computer to commit fraud is considered a crime of violence because of the potential harm to victims' financial well-being\\C. holding that threatening another with a weapon, even if not used, is a crime of violence for purposes of 18 U.S.C. § 16(b)\\D. holding that in terms of 18 U.S.C. § 16(b), verbal threats without the presence of a weapon do not qualify as a crime of violence\\E. holding that any crime which results in a financial penalty, irrespective of physical harm, is a crime of violence under 18 U.S.C. § 16(b)\\Answer: C
\end{tcolorbox}
    \caption{In-Context Example for CaseHOLD Generation}
    \label{fig:casehold_user_prompt}
\end{figure*}
%%%%%%%%%%%%%%%%%%%%%%%%%%%%%%%%%%%%%%%%%%%%%%%%%%%%%%%%%%%%%%%%%%%%%%

%%%%%%%%%%%%%%%%%%%%%%%%%%%%%%%%%%%%%%%%%%%%%%%%%%%%%%%%%%%%%%%%%%%%%%
\begin{figure*}[t!] 
    \centering
\begin{tcolorbox}[colback=gray!5!white,colframe=gray!75!black]
\textbf{System:} \\
You are TranscriptGPT, an AI agent who knows the intent of the transcript of different questions.\\You are training someone how to identify people's intents from what they have said.\\You are trying to give the user assistance by giving them more practice questions for the questions that they get wrong.\\\\Here are the requirements:\\1. A GPT language model should be able to complete the problem. For example, do not ask the assistant to create any visual or audio output. For another example, do not ask the assistant to wake you up at 5pm or set a reminder because it cannot perform any action.\\2. The question and options should be in english.\\3. The questions that you generate should have only 1 of the following intents:\\    - AddToPlaylist\\    - BookRestaurant\\    - GetWeather\\    - PlayMusic\\    - RateBook\\    - SearchCreativeWork\\    - SearchScreeningEvent\\4. The questions should always have 1 specific intent.\\5. The intent of the question must come from the list above.\\6. Don't make any mistakes with your answer yourself.\\7. Try not to copy too much information from the original problem. You don't want the user to just memorize the practice problems.\\8. Make the intent the same as the question that the user got wrong.\\9. The wrong answer choices can be any other reasonable answer, but it should be sufficiently different from the correct answer.\\10. The transcript should be something that an ASR model could output: it must sound like something a human could say.\\\\Always return your instructions in the form:\\1. Transcript: [CONTEXT]\\Intent: [INTENT]
\end{tcolorbox}
    \caption{System Prompt for SNIPS Generation}
    \label{fig:snips_system_prompt}
\end{figure*}
%%%%%%%%%%%%%%%%%%%%%%%%%%%%%%%%%%%%%%%%%%%%%%%%%%%%%%%%%%%%%%%%%%%%%%

%%%%%%%%%%%%%%%%%%%%%%%%%%%%%%%%%%%%%%%%%%%%%%%%%%%%%%%%%%%%%%%%%%%%%%
\begin{figure*}[t!] 
    \centering
\begin{tcolorbox}[colback=gray!5!white,colframe=gray!75!black]
\textbf{User:}
\\~\\
The following is a transcript whose intent the user got wrong including the correct answer from the answer key:\\\\The following is a transcript of something someone said.\\Classify the intent of the speaker into the following categories:\\- AddToPlaylist\\- BookRestaurant\\- GetWeather\\- PlayMusic\\- RateBook\\- SearchCreativeWork\\- SearchScreeningEvent\\\\Transcript: what s the weather here at 2 pm ',\\\\Intent: GetWeather\\\\Give me another 1 similar transcripts with the same answer (GetWeather).
\\~\\
\textbf{Assistant:}
\\~\\
Here's a similar transcript with the same intended answer category:\\\\1. Transcript: Will it rain in the city this afternoon?\\Intent: GetWeather
\end{tcolorbox}
    \caption{In-Context Example for SNIPS Generation}
    \label{fig:snips_user_prompt}
\end{figure*}
%%%%%%%%%%%%%%%%%%%%%%%%%%%%%%%%%%%%%%%%%%%%%%%%%%%%%%%%%%%%%%%%%%%%%%

%%%%%%%%%%%%%%%%%%%%%%%%%%%%%%%%%%%%%%%%%%%%%%%%%%%%%%%%%%%%%%%%%%%%%%
\begin{figure*}[t!] 
    \centering
\begin{tcolorbox}[colback=gray!5!white,colframe=gray!75!black]
\textbf{System:}
\\~\\
You are QuestionGPT, an AI agent who knows the class of different question.\\You are training someone how to classify different questions based on what the questions are asking form.\\You are trying to give the user assistance by giving them more practice questions for the questions that they get wrong.\\\\Here are the requirements:\\1. A GPT language model should be able to complete the problem. For example, do not ask the assistant to create any visual or audio output. For another example, do not ask the assistant to wake you up at 5pm or set a reminder because it cannot perform any action.\\2. The question should be in english.\\3. The questions that you generate should have only 1 of the following intents:\\    - ABBR (Abbreviation)\\    - ENTY (Entity)\\    - DESC (Description/Concept)\\    - HUM (Human)\\    - LOC (Location)\\    - NUM (Number)\\4. The questions should always have 1 specific class.\\5. The intent of the question must come from the list above.\\6. Don't make any mistakes with your answer yourself.\\7. Try not to copy too much information from the original problem. You don't want the user to just memorize the practice problems.\\8. Make the class the same as the question that the user got wrong.\\9. The question should be something that an ASR model could output: it must sound like something a human could say.\\\\Always return your instructions in the form:\\1. Question: [CONTEXT]\\Class: [INTENT]
\end{tcolorbox}
    \caption{System Prompt for TREC Generation}
    \label{fig:trec_system_prompt}
\end{figure*}
%%%%%%%%%%%%%%%%%%%%%%%%%%%%%%%%%%%%%%%%%%%%%%%%%%%%%%%%%%%%%%%%%%%%%%

%%%%%%%%%%%%%%%%%%%%%%%%%%%%%%%%%%%%%%%%%%%%%%%%%%%%%%%%%%%%%%%%%%%%%%
\begin{figure*}[t!] 
    \centering
\begin{tcolorbox}[colback=gray!5!white,colframe=gray!75!black]
\textbf{User:}
\\~\\
The following is a question that the user was unable to classify correctly:\\\\Classify the question into the following categories:\\- ABBR\\- ENTY\\- DESC\\- HUM\\- LOC\\- NUM\\\\Question: What country do the Galapagos Islands belong to ?\\Class: LOC\\\\Give me another 1 similar question with the same class (LOC).
\\~\\
\textbf{Assistant:}
\\~\\
Here's a similar question with the same class:\\\\1. Question: What country is the Great Barrier Reef located in?\\Class: LOC
\end{tcolorbox}
    \caption{In-Context Example for TREC Generation}
    \label{fig:trec_user_prompt}
\end{figure*}
%%%%%%%%%%%%%%%%%%%%%%%%%%%%%%%%%%%%%%%%%%%%%%%%%%%%%%%%%%%%%%%%%%%%%%

%%%%%%%%%%%%%%%%%%%%%%%%%%%%%%%%%%%%%%%%%%%%%%%%%%%%%%%%%%%%%%%%%%%%%%
\begin{figure*}[t!] 
    \centering
\begin{tcolorbox}[colback=gray!5!white,colframe=gray!75!black]
You are MovieReviewerGPT, an AI agent who knows how people review movies.\\You are training someone who is classifying movies as positive or negative.\\You are trying to give the user assistance by giving them more practice reviews for the reviews that they get wrong.\\\\Here are the requirements:\\1. A GPT language model should be able to complete the problem. For example, do not ask the assistant to create any visual or audio output. For another example, do not ask the assistant to wake you up at 5pm or set a reminder because it cannot perform any action.\\2. The review should be in english.\\3. The reviews that you generate should be either positive or negative.\\4. The reviews should always have 1 specific class.\\5. Don't make any mistakes with your answer yourself.\\6. Try not to copy too much information from the original problem. You don't want the user to just memorize the practice problems.\\7. Make the class the same as the review that the user got wrong.\\8. The review should be something that a user on the internet could type into a movie review forum.\\\\Always return your instructions in the form:\\1. Review: [CONTEXT]\\Sentiment: [SENTIMENT]
\end{tcolorbox}
    \caption{System Prompt for SST-2 Generation}
    \label{fig:sst2_system_prompt}
\end{figure*}
%%%%%%%%%%%%%%%%%%%%%%%%%%%%%%%%%%%%%%%%%%%%%%%%%%%%%%%%%%%%%%%%%%%%%%

%%%%%%%%%%%%%%%%%%%%%%%%%%%%%%%%%%%%%%%%%%%%%%%%%%%%%%%%%%%%%%%%%%%%%%
\begin{figure*}[t!] 
    \centering
\begin{tcolorbox}[colback=gray!5!white,colframe=gray!75!black]
\textbf{User:}
\\~\\
The following is a movie review that the user classified incorrectly including the correct classification: \\Classify the following movie review as positive or negative: as they come , already having been recycled more times than i 'd care to count \\Sentiment: negative\\\\Generate 1 more similar movie review with the same class.
\\~\\
\textbf{Assistant:}
\\~\\
Here's a similar question with the same class:\\\\1. Review: Feels like a reheated plot, utterly predictable and uninspired.\\Sentiment: negative
\end{tcolorbox}
    \caption{In-Context Example for SST-2 Generation}
    \label{fig:sst2_user_prompt}
\end{figure*}
%%%%%%%%%%%%%%%%%%%%%%%%%%%%%%%%%%%%%%%%%%%%%%%%%%%%%%%%%%%%%%%%%%%%%%

%%%%%%%%%%%%%%%%%%%%%%%%%%%%%%%%%%%%%%%%%%%%%%%%%%%%%%%%%%%%%%%%%%%%%%
\begin{figure*}[t!] 
    \centering
    \begin{tcolorbox}[colback=red!5!white,colframe=red!75!black]
    Seed Data Example: Katrina saw an ad that said if she signed up for her local recycling program, she could earn \$5.00. When she signed up, they told her for every friend that she referred, the friend would receive \$5.00 and she would receive another \$5.00 per friend. That day, she had 5 friends sign up and another 7 friends by the end of the week. How much money in total did she and her friends make?
    \end{tcolorbox}
    
    \begin{tcolorbox}[colback=blue!5!white,colframe=blue!75!black]
    GPT-4-Turbo: \textbf{Michael joined a library program that rewards \$4.00 for becoming a new member. Additionally, if he recommends the program to his friends, each friend gets \$4.00 and Michael also gets \$4.00 for each referee. Michael got 3 friends to sign up on the first day and another 6 friends joined the program over the next two weeks.} How much money in total did Michael and his friends earn?
    \end{tcolorbox}
    
    \begin{tcolorbox}[colback=green!5!white,colframe=green!75!black]
    GPT-3.5: Sophia saw an ad stating that if she signed up for her local fitness center, she could receive \$10 as a reward. \textbf{For every friend she refers, the friend receives \$10, and Sophia receives an additional \$10 per friend. On the first day, she referred 3 friends, and by the end of the week, she referred 6 more friends.} How much money did Sophia and her friends make in total?
    \end{tcolorbox}
    
    \begin{tcolorbox}[colback=yellow!5!white,colframe=yellow!75!black]
    Airoboros: Sophie saw an ad that said if she signed up for her local recycling program, she could earn \$3.00. When she signed up, they told her for every friend that she referred, the friend would receive \$3.00 and she would receive another \$3.00 per friend. That day, she had 7 friends sign up and another 9 friends by the end of the week. How much money in total did she and her friends make?
    \end{tcolorbox}
    
    \begin{tcolorbox}[colback=gray!5!white,colframe=gray!75!black]
    Llama-2-70B: \textbf{Joe signed up for his local recycling program and was promised \$10 for every friend he referred.} When he signed up, they told him for every friend that he referred, the friend would receive \$10 and he would receive another \$10 per friend. That day, he had 3 friends sign up and another 8 friends by the end of the week. How much money in total did he and his friends make?
    \end{tcolorbox}
    \caption{Example of new training data generated by various models for the same seed data example on GSM8K. The bold portions in each cell qualitatively highlight sentences that have been rewritten beyond simple noun or verb replacement. GPT models rewrite the seed data to a greater degree, while Airoboros and Llama-2-70B generally make fewer alterations to the sentence structure.}
    \label{fig:model_new_training_data}
\end{figure*}
%%%%%%%%%%%%%%%%%%%%%%%%%%%%%%%%%%%%%%%%%%%%%%%%%%%%%%%%%%%%%%%%%%%%%%

\end{document}